\documentclass[preprint,review,12pt]{elsarticle}
\usepackage[colorlinks,linkcolor=blue]{hyperref}

\usepackage{amssymb}

\usepackage{lineno}
\usepackage{multirow}
\usepackage{amsmath}
\usepackage{color}
\usepackage{float}
\usepackage{marvosym}
\journal{Pattern Recognition}

\begin{document}

\begin{frontmatter}



\title{FETNet: Feature Erasing and Transferring Network for Scene Text Removal}

\author[label1]{Guangtao Lyu}
\author[label1]{Kun Liu}
\author[label1]{Anna Zhu \corref{cor1} }

\author[label2]{Seiichi Uchida}
\author[label2]{Brian Kenji Iwana}
\address[label1]{School of Computer Science and Artificial Intelligence, Wuhan University of Technology, Wuhan, China}

\address[label2]{Human Interface Laboratory, Kyushu University, Fukuoka, Japan}

\cortext[cor1]{Corresponding author. \href{mailto:annazhu@whut.edu.cn}{annazhu@whut.edu.cn}}

\begin{abstract}
   The scene text removal (STR) task aims to remove text regions and recover the background smoothly in images for private information protection. Most existing STR methods adopt encoder-decoder-based CNNs, with direct copies of the features in the skip connections. However, the encoded features contain both text texture and structure information. The insufficient utilization of text features hampers the performance of background reconstruction in text removal regions. To tackle these problems, we propose a novel Feature Erasing and Transferring (FET) mechanism to reconfigure the encoded features for STR in this paper. In FET, a Feature Erasing Module (FEM) is designed to erase text features. An attention module is responsible for generating the feature similarity guidance. The Feature Transferring Module (FTM) is introduced to transfer the corresponding features in different layers based on the attention guidance. With this mechanism, a one-stage, end-to-end trainable network called FETNet is constructed for scene text removal. In addition, to facilitate research on both scene text removal and segmentation tasks, we introduce a novel dataset, Flickr-ST, with multi-category annotations. A sufficient number of experiments and ablation studies are conducted on the public datasets and Flickr-ST. Our proposed method achieves state-of-the-art performance using most metrics, with remarkably higher quality scene text removal results. The source code of our work is available at: \href{https://github.com/GuangtaoLyu/FETNet}{https://github.com/GuangtaoLyu/FETNet.}

\end{abstract}


%


\end{frontmatter}



\section{Introduction}
\label{sec:intro}

Natural scene images contain quite a lot of sensitive and private information, such as names, addresses, and cellphone numbers. The developed OCR technics \cite{m_Nguyen-etal-CVPR21} can extract those texts automatically and easily. To prevent the leakage of private text information in images being used illegally, the scene text removal \cite{2017Scene} (STR) task has been proposed recently.

STR aims to automatically erase text regions and fill them with background information in natural images. Recently, it has attracted significant attention in the computer vision community due to its valuable applications in privacy protection \cite{2014Selective}, visual translation \cite{2021TextOCR}, information reconstruction, \cite{2020Textual} and image editing \cite{2020SwapText}.

\begin{figure}[ht]
\centering
  \includegraphics[width=0.7\textwidth]{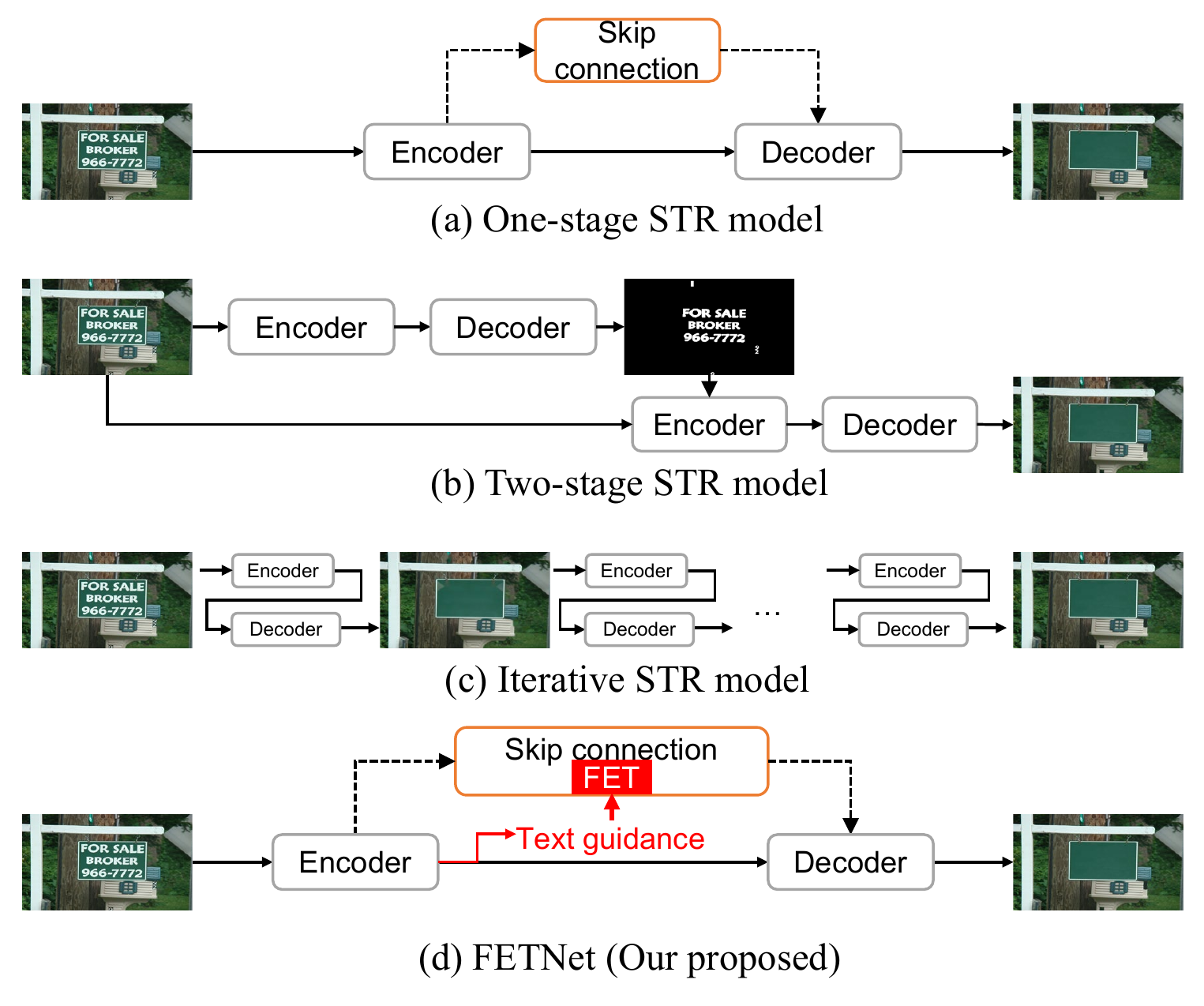}
  \caption{Comparison of three kinds of STR frameworks and our proposed FETNet. The one-stage based FETNet reconfigures features in skip-connections by using text guidance. It takes the advantage of lightweight characteristics of the one-stage STR model and the controllable characteristics of multi-stage STR models.}
  \label{fig_model}
\end{figure}

Generally, the existing STR research can be classified into three categories: one-stage, two-stage, and iterative-based methods, as displayed in Fig.\ref{fig_model}. One-stage models \cite{2017Scene,2018EnsNet} are built on single encoder-decoder architectures with the scene text image as input and directly outputting the text-removed result. These models are usually lightweight and run fast. However, since they learn both text detection and inpainting implicitly within a single network, the text localization mechanism is weakened and makes the text erasing process uncontrollable.

Two-stage STR methods adopt two or more encoder-decoders. They detect scene text in the first step and the inpaint the detected region as a part of the background in the second step~\cite{2019MTRNet,Zdenek2020WACV}. Alternatively, some two-stage methods~\cite{2021Detecting,2020EraseNet} use a coarse-to-fine strategy to obtain coarse removal results in the first stage and refine the removal performance in the second stage. Iterative-based STR methods~\cite{PERT} also use text detection results as image inpainting masks but perform multi-stage text erasure in an iterative way~\cite{2019Progressive}. Those methods are limited due to high reliability on text detection or coarse removal results front-end. Compared to the one-stage approach, they are inefficient and complex since all networks in different stages need to be trained. The above analysis makes us rethink the STR task: is it possible to adapt the architecture of the one-stage STR method since it is efficient, meanwhile integrating text controlling process to overcome its drawback?

In this paper, we propose an encoder-decoder-based convolutional neural network (CNN) called Feature Erasing and Transferring Network (FETNet), which uses text segmentation as guidance and fuses it for background inpainting in a single network. It belongs to the one-stage STR category. The encoder is shared by text segmentation and background inpainting sub-tasks, and the intermediate text segmentation result is reused in the decoder for text removal, forcing the two sub-tasks to cooperate with each other and also making the network training to be end-to-end.

The key idea of our FETNet is to utilize and process the different features extracted in different CNN layers for STR. CNNs for text segmentation generate different features in different layers \cite{2017Scene} (i.e., low-level features in shallow layers and high-level semantics in deep layers). In other words, the features from the shared encoder contain spatial text information in shallow layers and structural text information in deeper layers. Therefore, we design the {\em Feature Erasing and Transferring} (FET) mechanism containing three modules to process different information in each layer.

One key point is how to remove the various text information in the feature spaces. We propose to use Feature Erasing Modules (FEM), which is the first module in FET to remove text information. The STR task requires the text regions to be covered by smooth background. Therefore, the second key point is how to reconstruct the features in text regions and force the features to be similar to text-surrounded background features. To solve it, the {\em Attention Modules} (AM) and the {\em Feature Transfer Modules} (FTM) are sequentially used in FET to obtain background feature similarity, and then transfer the similar features. The FET mechanism is inserted between the encoder and decoder. We expect that if text information is successfully erased and refilled by background features, the output images after deconvolutions contain no text and only background features \cite{2018Generative,2020Rethinking}.

We conducted experiments on three datasets: the SCUT-EnsText \cite{2020EraseNet}, SCUT-Syn \cite{2018EnsNet}, and Flickr-ST datasets \cite{2017Scene}. Among them, the SCUT-EnsText dataset only contains text-removed ground truth (GT) and text bounding boxes in images, which may not be precise when recovering background textures around the text. SCUT-Syn is a synthetic dataset with fake scene texts attached randomly in the images, which cannot reflect the real scenes. The Flickr-ST dataset is a scene text dataset \cite{2017Scene} with pixel-level character segmentation and category labels. To better utilize it for the STR task, we carefully erase each character and inpainting its background for all 3,004 images. This dataset can also be used for scene text detection, text recognition, text segmentation, etc. 

The experimental results on the three datasets demonstrate that FETNet can outperform previous state-of-the-art (SOTA) STR methods both qualitatively and quantitatively. In addition, the proposed FET mechanism can plug in other two-stage STR models for performance enhancement.

We summarize the contributions of this work as follows: (1) We propose a novel FETNet which could remove scene text near-completely in images. It fully utilizes CNN features through a designed Feature Erasing and Transferring (FET) mechanism, which could erase and reconfigure text features through three cascaded modules. This mechanism can be plugged into other two-stage STR models for performance enhancement. (2) Our method is formulated in a one-stage way and is trained in an end-to-end manner. It uses the intermediate text segmentation features as guidance to control the feature erasing regions in the decoder, which overcomes the drawback of one-stage STR methods and realizes computationally efficient STR. (3) We introduce a novel Flickr-ST dataset with multi-category careful annotations, which can benefit our work and various scene text-related research. (4) Our FETNet achieves remarkable quantitative and qualitative results on public synthetic and real scene datasets. Ablation studies illustrate the proposed architecture and its components play important roles in scene text removal.

\section{Related Works}

\textbf{Scene Text Segmentation.} Scene text segmentation aims to detect text at the pixel level, i.e., extracting all text pixels of an image. Traditional methods used thresholding-based \cite{2014Text} and edge detection-based \cite{2013Multiscale} image processing techniques for text segmentation on document images. However, they have performed unsatisfactorily on scene text images due to the varying text appearances and complex arrangements. Other approaches employed machine learning algorithms, e.g., Markov Random Field (MRF) \cite{2015MRF} and Conditional random field (CRF) \cite{2010Conditional} to bipartite scene text images. But, they missed a lot of text regions and detected false positives in complex backgrounds. Maximally Stable External Regions (MSER) \cite{2014Scene} applied both low and middle-level features to extract text candidates with text-specific features to improve the segmentation performance. But, it cannot be trained end-to-end and the features were hand-crafted.

In recent years, deep learning-based text segmentation methods \cite{2017Scene,2020Weak,2021Rethinking,2021TIP} were proposed and showed great superiority. Tang et al. \cite{2017Scene} first proposed a CNN-based coarse-to-fine pixel-level text segmentation model. SMANet \cite{2020Weak} adopted the encoder-decoder structure from PSPNet \cite{2017Pyramid} and designed a multi-scale attention module for text segmentation. TextRNet \cite{2021Rethinking} was a text-specific network that adapted to the unique text properties. Wang et al. \cite{2021TIP} proposed a mutually guided network to produce polygon-level mask and pixel-level text mask in two branches and trained it semi-supervised.

\textbf{Image Inpainting.}
Traditional image inpainting methods adopted diffusion techniques \cite{2001Filling} or patch-matching \cite{2004Region,2009Patchmatch}, to fill in the missing regions by neighborhood appearances or similarity patches. However, they are unable to generate semantic content.

Deep learning-based inpainting approaches have recently been proposed, which can generate meaningful content for filling in the missing regions. Context Encoder \cite{2016Context} was a pioneering work, which utilized adversarial learning for feature extraction and reconstruction. Afterward, many researchers \cite{2018Generative,2020Recurrent} adopted attention mechanisms for generating the missing content, e.g., CAM \cite{2018Generative}, KCA \cite{2020Recurrent}, etc.

In addition, several studies \cite{2018Image,2019Free} modified the regular convolution operations to overcome the weakness of the vanilla convolution, e.g., partial convolution \cite{2018Image}, gated convolution \cite{2019Free}, etc., for image inpainting.

\textbf{Text Removal.} Text removal studies can be classified into three categories: one-stage, two-stage, and iterative-based methods.

One-stage STR methods treat the task as an image translation problem. For instance, the well-known Pix2Pix \cite{2017Image} uses patch-GAN to learn the mapping from
input to output and can be applied for scene text removal. Similarly, Scene Text Eraser (STE)~\cite{2017Scene} provided an STR baseline by constructing an encoder-decoder architecture for erasing text in patches of images. But, only using patches extracted from single-scaled sliding-windows lost the global context information, failing for large text region removal. EnsNet \cite{2018EnsNet} adopted cGAN \cite{2017Image} with several designed loss functions and a novel lateral connection to further enhance the STR performance.

Two-stage STR methods either adopt text detection-to-image inpainting flow or coarse-to-fine pipeline. MTRNet \cite{2019MTRNet} applied a text-awareness auxiliary mask for a Pix2Pix-based model, which can focus on the task of text inpainting. However, the implicit text region guidance may cause excessive erasure to non-text regions. To overcome this problem, Tang et al.~\cite{StrokeErase2021tang} adopted a pixel-level mask of text stroke in the second stage to explicitly guide text inpainting. EraseNet~\cite{2020EraseNet} used a coarse-to-fine procedure, decomposing scene text image to text segmentation and coarse text removal heads in the coarse stage and then refining the coarse results in the second stage. Differently, MTRNet++~\cite{tursun2020mtrnet++} and Cho et al.~\cite{2021Detecting} constructed the refinement network inputting both the coarse output and text mask from the first erasure stage. CTRNet \cite{liu2022don} developed a Local-global Content Modeling (LGCM) block to provide both low-level and high-level contextual guidance, which could effectively enhance the performance of texture restoration for complex backgrounds. Bian et al.~\cite{bian2022scene} proposed a cascaded GAN-based model by decoupling text removal into text stroke detection and stroke removal.

Iterative STR models performed the erasure several times. PERT~\cite{2019Progressive} used balanced multi-stage erasure with several progressive erasing with explicit text region guidance. PSSTRNet~\cite{lyu2022psstrnet} proposed a new mask update module to obtain increasingly accurate text masks and adopted an adaptive fusion strategy to take full advantage of the results of different iterations. Our method belonged to the one-stage category but utilized text segmentation information to make the erasure controllability.



\section{Proposed Method}

\subsection{Overall pipeline}

\begin{figure*}[b]
  \includegraphics[width=\textwidth]{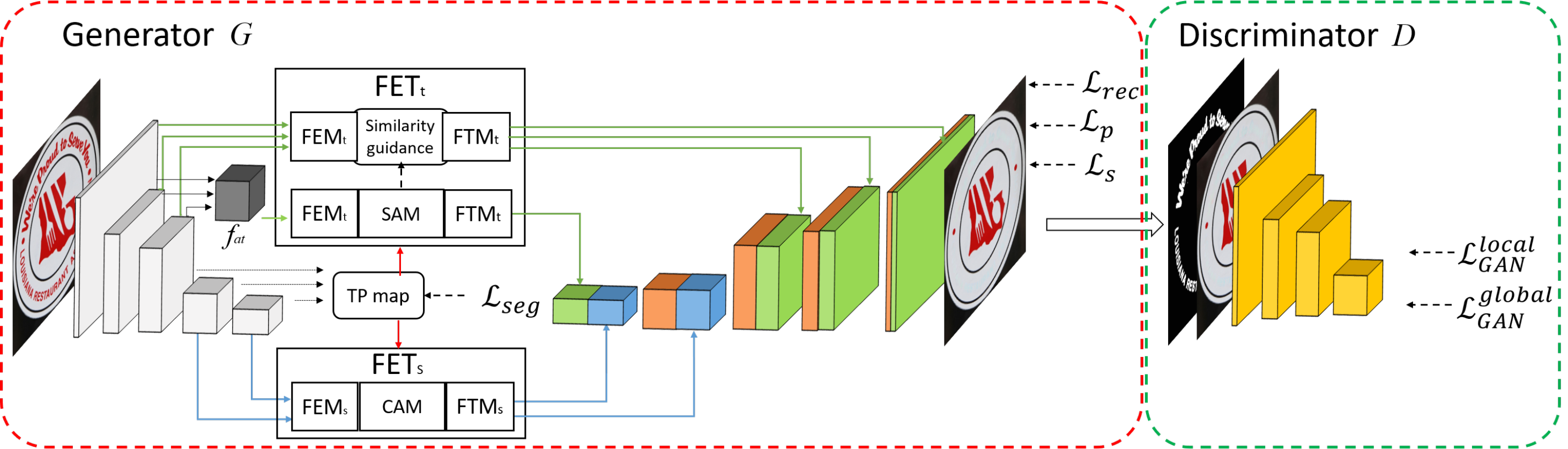}
  \caption{ The overall architecture of FETNet.}
  \label{model}
\end{figure*}

As shown in Fig.\ref{model}, the pipeline of our model consists of two parts: a generator $G$ and a discriminator $D$. In $G$, one simple encoder-decoder structure \cite{2020Rethinking} is constructed, consisting of five residual convolutional layers in the encoder and five residual convolutional (3$\times$3) layers in the decoder, respectively. Following the resnet backbone architecture design, the first to fifth layers of the encoder has kernel sizes of 7, 5, 3, 3, and 3 respectively. And the output feature maps of them have the size of $1, \frac{1}{2}, \frac{1}{4}, \frac{1}{8}, \frac{1}{16}$ of the input image, respectively. In $D$, a four-layer CNN is designed to judge images as real or fake globally and locally.

Since one-stage STR methods lack explicit text region guidance, we extend a branch for text segmentation. In the CNN, the color, edge, and texture features are extracted in shallow layers, and the deep layers contain more structured semantic features (as shown in Fig.\ref{visual}). The text segmentation results only contain the pixel-level text location but no texture. Therefore, we only extract features from the last three layers in the encoder for text segmentation. Three additional deconvolution layers are built to connect the encoders as a U-Net structure, followed by a sigmoid function to obtain the text confidence map $C_t$. $C_t$ is unsampled or downsampled to different scales according to the different feature sizes.

The FET mechanism is designed and inserted between the encoder and decoder to reconfigure the texture and structure features through text feature deletion and  background feature filling. It works in two different ways, i.e., texture FET (FET$_t$) and structure FET (FET$_s$), in different layers. Since the encoder jointly learns the structure and texture features of text and background, directly concatenating those features \cite{2020EraseNet} in skip-connections will involve text information. This will cause unclean text removal results. Additionally, the text in the shallow layers of CNNs has more local and texture information, and they contain more abstract and structural information channel-wisely in deep layers.
Therefore, we reconfigure the features from the last two layers in the encoder as structure features. Meanwhile, we reorganize the first three-layer features as texture features. FET$_t$ and FET$_s$ are responsible for these two types of features, respectively. Both are comprised of FEM, FTM, and one attention module, reorganizing the features in different ways.

The structure features from the last layer and the texture features from the first three layers are reorganized through the FET$_s$ and FET$_t$, respectively. Then, they are concatenated as the initial global features in the decoder. Instead of using normal skip-connections to maintain background information, we insert the FET into it. The features from each encoder layer go through FET and are then concatenated with corresponding deconvolutional decoder features. The final output is the text removal images. In the following subsections, we present the details of the FET$_t$ and FET$_s$ and introduce the loss functions for training this one-stage FETNet.

\subsection{Texture FET mechanism}\label{sect1}

\begin{figure}
\centering
  \includegraphics[width=0.8\linewidth]{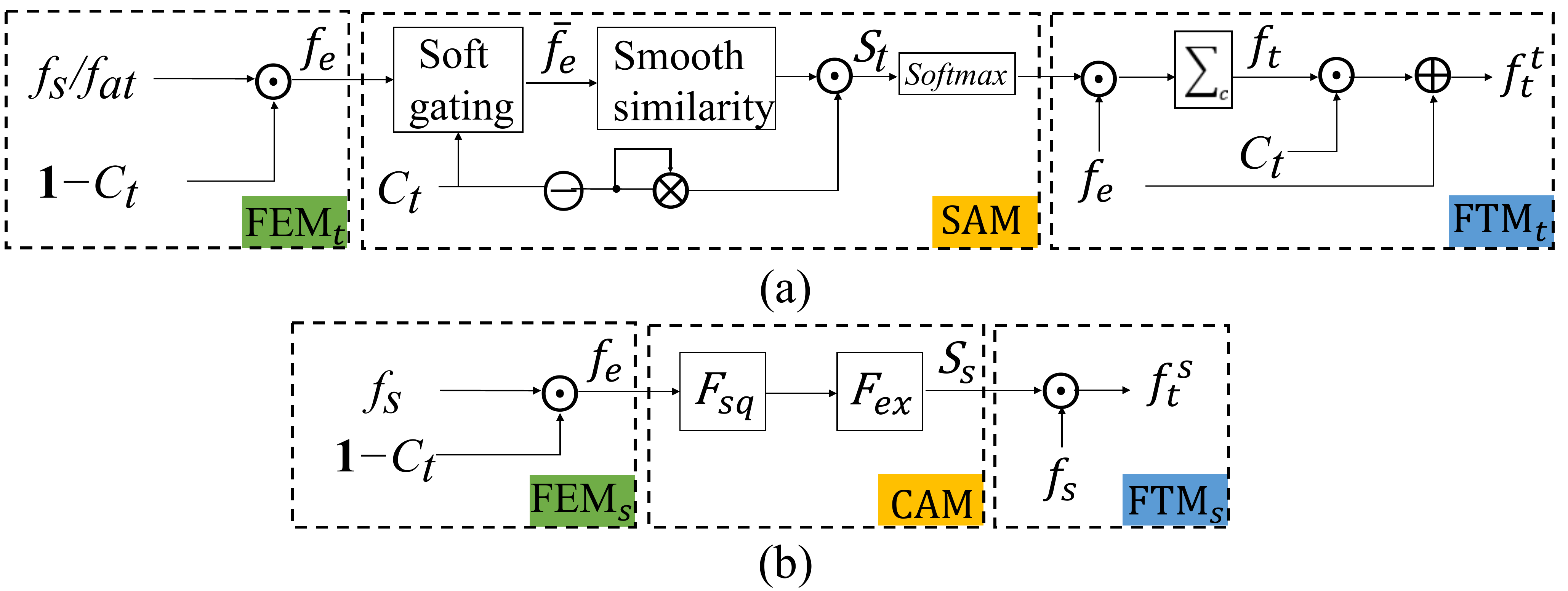}
  \caption{ The texture FET mechanism (a) and the structure FET mechanism (b).}
  \label{fet_block}
\end{figure}

As shown in Fig.\ref{fet_block}(a),  
FET$_t$ contains a texture feature erasing module (FEM$_t$), a spatial attention module (SAM), and a texture feature transferring module (FTM$_t$). The FEM$_t$ is designed to remove text features from the shallow layers with the guidance of $C_t$. In order to refill corresponding correct background features in erased regions, a spatial attention operation is implemented to get feature similarity in each position. Then, the FTM$_t$ is used to transfer the correct background texture features to text regions. By using the three modules successively, FET$_t$ can substitute the text features with consistent background features.

The input to FEM$_t$ is aggregated features from different layers. We use multi-scale aggregation \cite{2016Multi} to aggregate texture features $f_{at}$ through resizing and transforming the CNN feature maps $f_s$$\in$${\rm\mathbb{R}}^{w_s\times h_s\times c_s}$ from three shallow layers to the same size and concatenating them accordingly. The aggregated texture features $f_{at}$ are then input to FEM$_t$ to get text erased feature $f_{e}$ as in Eq. \eqref{eq1}, where $\odot$ refers to Hadamard product.
\begin{equation}
 f_{e}=(1-C_t)\odot f_{at}.
  \label{eq1}
\end{equation}
In SAM, three soft gating convolutions~\cite{2019Free} expressed in Eq.~\eqref{eq2} are used to fill the erased feature with rough features which are dynamically selected from other channels and spatial locations:
\begin{equation}
\begin{aligned}
f_g=W_g\ast [f_e, C_t],\\
f_d=W_f\ast [f_e, C_t],\\
\bar{f}_{e}=\phi(f_d)\odot\sigma(f_g),
\end{aligned}
  \label{eq2}
\end{equation}
\noindent where $\ast$ represents the convolution operation and [$\cdot$,$\cdot$] denotes concatenation. $f_g$$\in$${\rm\mathbb{R}}^{w_{e}\times h_{e}\times 1}$ is the gating features and becomes gating values between zeros and ones after sigmoid function $\sigma$. $f_d$$\in$${\rm\mathbb{R}}^{w_{e}\times h_{e}\times c_{e}}$ is transformed features fusing text information. $\phi$ can be any activation function. We select the ReLU function for our method. $\bar{f}_{e}$ is the coarse filling features. $W_g$ and $W_f$ are two different trainable convolutional filters.

Directly using $\bar{f_{e}}$ will generate very blur results. Therefore, it is only represented as coarse features. The spatial attention searches for possible textures in the background and uses them to replace the textures in regions with a high probability of being text. We measure the cosine similarity between features in each pair of locations as Eq.\eqref{eq3}. 
\begin{equation}
 \bar{S}_{(x_i,y_i),(x_j,y_j)}=\left\langle\frac{\bar{f_{e}}(x_i,y_i)}{\|\bar{f_{e}}(x_i,y_i)\|},\frac{\bar{f_{e}}(x_j,y_j)}{\|\bar{f_{e}}(x_j,y_j)\|}\right\rangle.
  \label{eq3}
\end{equation}

Therefore, we could get a smooth attention map $S$$\in$${\rm\mathbb{R}}^{w_{e}h_{e}\times w_{e}h_{e}}$. However, the attention of high probability text position may also focus on other text positions. Our goal is to transfer background features onto only the erased text regions, so spatial background concurrency probability should also be considered. The background concurrency attention $S_t$$\in$${\rm\mathbb{R}}^{w_{e}h_{e}\times w_{e}h_{e}}$ is written as Eq. \eqref{eq5}, where $\otimes$ is the Kronecker product.
\begin{equation}
S_t=(1-C_t)\otimes(1-C_t)^T\odot S.
  \label{eq5}
\end{equation}

\noindent$S_t$ is reshaped to $w_{e}$$\times$ $h_{e}$$\times$$w_{e}$$h_{e}$, and Softmax function is performed on each channel respectively. The output of ($i\times w_e$+$j$)$_{th}$ channel denoting as Softmax$_c$($S_t$)$_{i,j}$ represents the global background attention weights for position ($i$, $j$).

Then, FTM$_t$ calculates the transferred texture features $f_t$($i$,$j$) for each position ($i$, $j$) as Eq. \eqref{eq6}. We expect only background features to be transferred to erased text regions. The features of the other background regions should not be changed. So, we update the transferred texture features considering the text probability as in Eq. \eqref{eq7}.
\begin{equation}
f_t(i,j)=\sum(\mathrm{Softmax}_c(S_t)_{i,j}\odot f_e).
  \label{eq6}
\end{equation}
\begin{equation}
f_t^t=C_t\odot f_t+f_e.
  \label{eq7}
\end{equation}

For the layer-wise texture features $f_s$ transferring in skip-connections, we do not recompute spatial attention but share the rescaled attention weights Softmax$_c$($S_t$) as the similarity guidance. Features in each of the first three layers go through FEM$_t$ to get $f_e$, and then FTM$_t$ to get $f_t^t$.

\subsection{Structure FET mechanism}\label{sect2}
Similarly, FET$_s$ contains a FEM$_s$, FTM$_s$, and a channel attention module (CAM). The purpose of this mechanism is to enhance the structure features of the background but suppress the
structure features of texts. Since features in each channel of deep layers contain more global information, we follow the squeeze and excitation operation \cite{2018Squeeze} to get the background attention of each channel. The features flowing in FET$_s$ are displayed in Fig.\ref{fet_block}(b).

For features $f_s$ in layer $s_{th}$ $\in$[4,5], FEM$_s$ first erases the text features by element-wise multiplication with 1-$C_t$ as Eq. \eqref{eq1}. Then, squeeze and excitation operation is employed for feature $f_e$, to get the channel attention $S_c$$\in$${\rm\mathbb{R}}^{1\times 1\times c_s}$ in CAM, as written in Eq. \eqref{eq8}. $F_{sq}$ is to squeeze features, which is
achieved by using global average pooling to generate channel-wise statistics. $F_{ex}$ is used to excite channel attention by forming two fully-connected layers and a sigmoid function. The scores in $S_c$ express the importance of background structure features in the corresponding channels.

Finally, FTM$_s$ implements channel-wise multiplication between the channel attention score $S_s$ and the original features $f_s$ to output the transferred structure features $f_t^s$.
\begin{equation}
 S_c=F_{ex}(F_{sq}((f_{e})))
  \label{eq8}
\end{equation}
\begin{equation}
 f_t^s=S_c\odot f_s.
  \label{eq9}
\end{equation}
\subsection{Loss functions}\label{sect3}
We introduce several loss functions for FETNet learning, including pixel reconstruction loss, perceptual loss, style loss, and relativistic average LS adversarial loss. Given the input scene text image $I_{i}$, text-removed GT $I_{gt}$ and the binary text GT mask $M_{gt}$ (0 for non-text regions and 1 for text regions in $M_{gt}$), the STR output from FETNet is denoted as $I_{o}$ and text segmentation results as $M_{o}$. $M_{o}$ is obtained by thresholding on $C_t$.

\textbf {Pixel Reconstruction Loss.} First, we impose a reconstruction constraint that forces both backgrounds and removed text regions to approach the real GT $I_{gt}$, using a standard autoencoder L$_1$ loss:
\begin{equation}
    L_{rec} = ||(1 - M_{gt})\odot (I_{o} - I_{gt})||_1+\lambda_t||M_{gt}\odot (I_{o} - I_{gt})||_1,
    \label{eq10}
\end{equation}
\noindent where $\lambda_t$ is set to 5, which emphasizes the learning importance of text removed regions.

\textbf {Perceptual Loss.} To capture the high-level semantics and simulate human
perception of images quality, we utilize the perceptual loss \cite{johnson2016perceptual}:
\begin{equation}
\begin{split}
	L_{p} = \sum_{i}\frac{1}{N_i}||\Phi_i(I_{o}) - \Phi_i(I_{gt})||_1 + \sum_{i}\frac{1}{N_i}||\Phi_i(I_{c}) - \Phi_i(I_{gt})||_1,
\end{split}
\label{eq11}
\end{equation}
\begin{equation}
I_{c} = I_{i}\odot (1-M_{gt}) + I_{o}\odot M_{gt},
\label{eq12}
\end{equation}

\noindent where $\Phi_i$ is the activation map of the $i$-th layer of the VGG-16 backbone. $I_{c}$ is the composted image with background of original input $I_{i}$ and text removed regions of $I_{o}$.

\textbf {Style Loss.} We compute the style loss \cite{2016Image} as in Eq.\eqref{eq13}, where $G_i$($\Phi$) is the Gram matrix constructed from the selected activation maps.
\begin{equation}
\begin{split}
	L_{s} = ||G_i(\Phi(I_{o})) - G_i(\Phi(I_{gt}))||_1 +||G_i(\Phi(I_{o})) - G_i(\Phi(I_{gt}))||_1.
\end{split}
\label{eq13}
\end{equation}

\textbf {Segmentation Loss.} For learning of text segmentation module, we use dice loss  \cite{milletari2016fully} as Eq. \eqref{eq14}, which considers the contour similarity between the prediction results in $M_{o}$ and GT $M_{gt}$.
\begin{equation}
	L_{seg} =  1-\frac{2\sum_{x,y}(M_{o}(x,y)\times M_{gt}(x,y))}{\sum_{x,y}(M_{o}(x,y)^2 + M_{gt}(x,y)^2)}.
\label{eq14}
\end{equation}

\textbf {Adversarial Loss}. We utilize the global and local discriminators \cite{2019Free} for perception enhancement. The adversarial loss of $G$ and $D$ is defined as Eq.\eqref{eq15} and Eq.\eqref{eq16}, respectively.
\begin{equation}
	L_G^{adv} = -\mathbb{E}[\mathrm{log}D(G(I_{i}))],
\label{eq15}
\end{equation}
\begin{equation}
	L_D^{adv} =  \mathbb{E}[\mathrm{log}D(I_{gt})]+\mathbb{E}[\mathrm{log}(1-D(G(I_{i})))].
\label{eq16}
\end{equation}

The total loss for training the generator in FETNet is the weighted combination of the losses mentioned above:
\begin{equation}
\begin{split}
	L_{total} = L_{rec}+\lambda_sL_{s}+\lambda_pL_{p}+\lambda_mL_{seg}+\lambda_gL_G^{adv}.
\end{split}
\end{equation}
\noindent In experiments, we set $\lambda_s$, $\lambda_{p}$, $\lambda_{m}$ and $\lambda_g$ to 60, 0.05, 1.5, 0.05 respectively, basing on the their effect for training the model~\cite{2018Image, 2018EnsNet}. 

\section{Datasets}

\begin{figure}[t]
  \includegraphics[width=\linewidth]{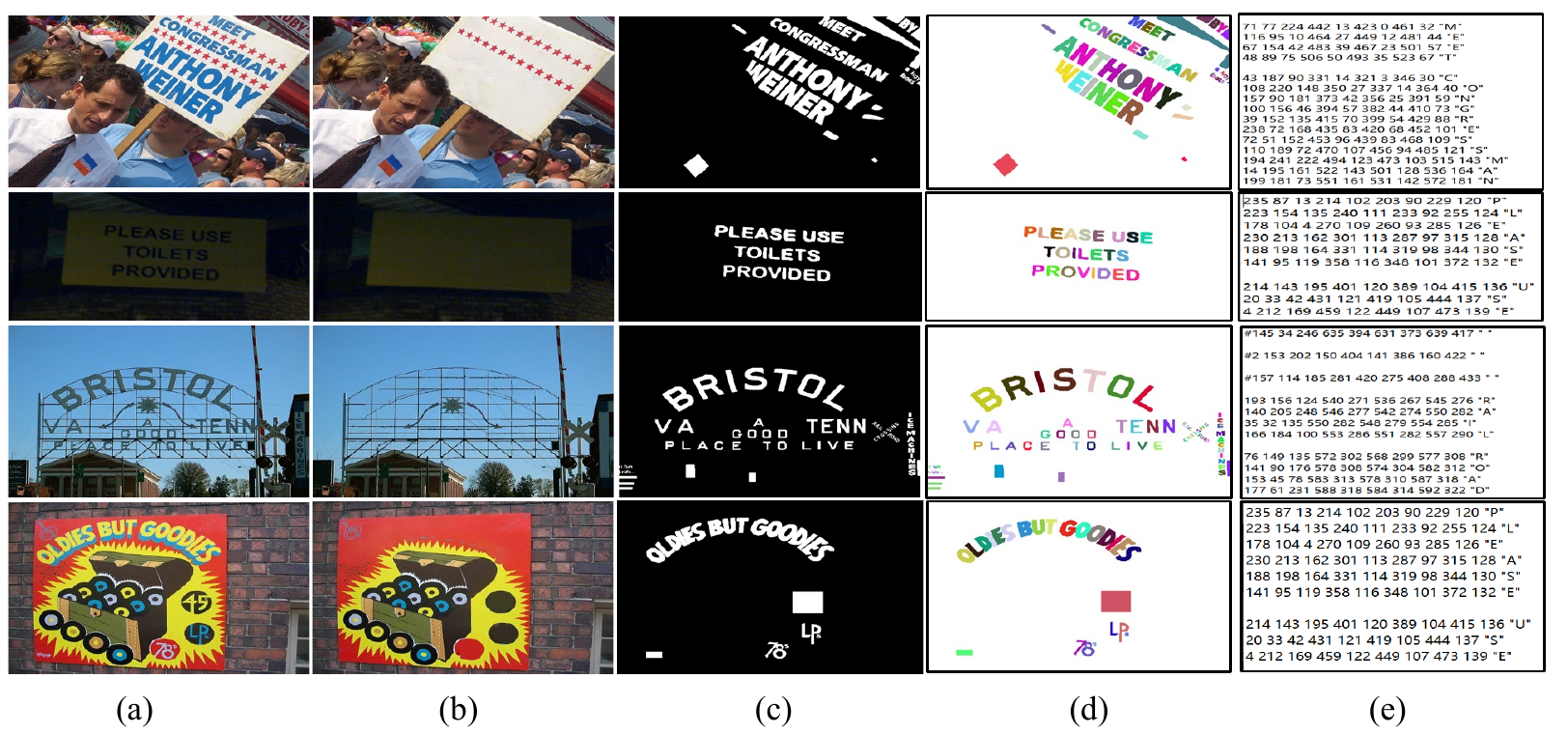}
  \caption{Changeling examples and the corresponding annotations of the Flickr-ST dataset. (a) text image (b) text-removed image (c) pixel-level text mask (d) character-wise pixel-level mask (e) part of character-wise bounding polygons}
  \label{flickr_hard}
\end{figure}

\textbf{Flickr-ST}. The Flickr-ST dataset is a real-world dataset \cite{2017Scene} including 3,004 images with 2,204 images for training and 800 images for testing. The scene text in Flickr-ST has arbitrary orientations and shapes. Some examples are shown in Fig.\ref{flickr_hard}(a). It provides five types of annotations, text removed images (in Fig.\ref{flickr_hard}(b)), pixel-level text masks (in Fig.\ref{flickr_hard}(c)), character instance segmentation labels, category labels, and character-level bounding box labels(in Fig.\ref{flickr_hard}(d) and (e)). The word-level scene text regions can be calculated implicitly from those labels. To the best of our knowledge, Flickr-ST is the only dataset with such comprehensive annotations for scene text related tasks.

The character-level segmentation labels, bounding box labels, and category labels are stored in .png and .xml file formation for each image as shown in Fig.\ref{flickr_hard}(d) and (e) respectively. In the .png file, each character in the corresponding original scene text image is annotated with distinguished color. Each line in the .xml file contains nine numbers and one string. The numbers represent the RGB value and bounding box coordinates of the labeled character in the .png file. The string denotes the character class. Additionally, the blank line interval represents the end of a word-level text. Thus, it is very simple to obtain the GT of word-level regions.

There are some non-readable or too-small characters that are very difficult to label. We consider them as do-not-care characters. Therefore, they are labeled by quadrilateral regions in .png file and have no corresponding character class in .xml file.

Text-removed images are generated from original images with scene text erased and filled with a visually plausible background to maintain the consistency of erased text regions and their surrounding texture. All the filling and erasing operations are constructed using Adobe Photoshop. We mainly use Spot Heading Brush and Pattern Stamp functions. Then, we merge the text-removed image and the corresponding text image with the text masks to modify the background area as little as possible.

Pixel-level text mask is a binary image to distinguish text and non-text. We ignore some text effects like shadow and decoration during labeling.

\textbf{SCUT-Syn}. This synthetic dataset only includes English text instances, with 8,000 images for training and 800 images for testing. The details for generating and using these synthetic text images can be found in \cite{gupta2016synthetic}.

\textbf{SCUT-EnsText}. It contains 2,749 training images and 813 test images which are collected in real scenes with diverse text included. More descriptions about this dataset refer to the work  \cite{2020EraseNet}.

\textbf{Evaluation Metrics.} A text detector CRAFT \cite{2019Character} is employed to detect text on the output STR images by measuring the precision, recall, and F-score. The lower their values, the better text removal performance. Six image-level evaluation
metrics are also adopted for measurement of the generated STR images with GT images, i.e., Peak Signal to Noise Ratio (PSNR), Mean Square Error (MSE), Mean Structural Similarity (MSSIM), Average Gray-level Error (AGE), percentage of Error Pixels (pEPs), and percentage of Clustered Error Pixels (pCEPS) \cite{2018EnsNet}. A higher MSSIM and PSNR, and a lower AGE, pEPs, pCEPS, and MSE indicate better results.

\section{Experiments and Results}

\subsection{Implementation details}
We train FETNet on the training sets of SCUT-EnsText, SCUT-Syn, and Flickr-ST and evaluate them on their corresponding testing sets. The masks are generated by subtracting the input images and the corresponding STR GT, i.e., text removal images. We follow  \cite{2020EraseNet} to apply data augmentation during training. The model is optimized using the Adam algorithm. The initial learning rate of the generator is set to 0.001, and the discriminator is 0.002. Following the training procedures of GAN, we alternately train the generator and discriminator for 500 epochs in a single NVIDIA GPU with a batch size of 6 and input image size of 256$\times$256.

\subsection{Ablation study}

\subsubsection{The effect of different components in FETNet}

\begin{table}
  \caption{Ablation study results of different modules effect on SCUT-Syn.}
  \center
  \resizebox{0.7\columnwidth}{!}{
  \begin{tabular}{|c|c|c|c|c|c|c|}
    \hline Method&PSNR $\uparrow$ & MSSIM $\uparrow$ & MSE $\downarrow$ & AGE $\downarrow$ & pEPs $\downarrow$ & pCEPs $\downarrow$\\
    \hline
    - FEM &34.23 & 96.24 & 0.0010 & 3.0373 & 0.0175 & 0.0074 \\
    \hline
    - FTM &37.07 & 97.29 & 0.0004 & 2.3667 & 0.0063 & 0.0014 \\
    \hline
    - Simlarity &37.95 & 96.92 & 0.0002 & 1.3293 & 0.0067 & 0.0011 \\
    \hline
    + Output Mask& 36.21 & 96.62 & 0.0006 & 2.5429 & 0.0107 & 0.0038 \\
    \hline
    Full model & \textbf{39.14} & \textbf{97.97} & \textbf{0.0002} & \textbf{1.2608} & \textbf{0.0046} & \textbf{0.0008} \\
    \hline
\end{tabular}}
\label{tab1}
\end{table}

\begin{figure}
  \includegraphics[width=\linewidth]{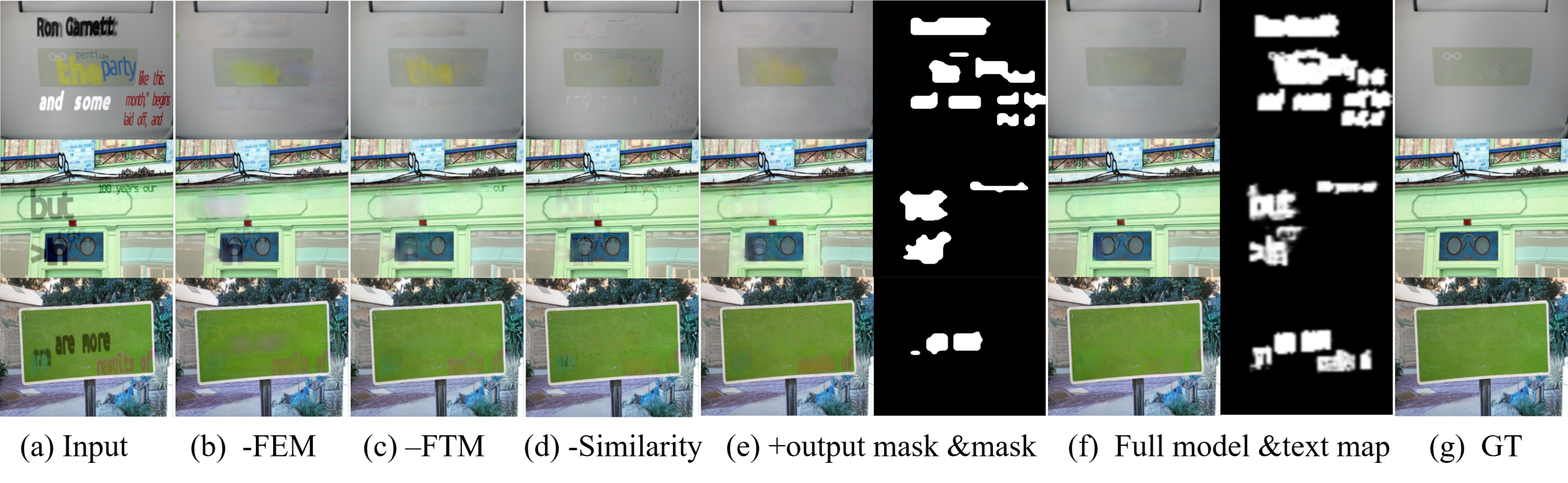}
  \caption{Ablation study on different models. (a) Input image. (b)(c)(d)(e)(f) show the STR results by removing FEM, removing FTM, removing Similarity, using generated mask instead of a text confidence map, and the full model. (g) is the STR GT.}
  \label{fig3}
\end{figure}

In this section, we verify the effect of different components of FETNet on the SCUT-Syn dataset. In total, we set five comparison experiments as follows: removing FEM ($-$FEM), removing FTM ($-$FTM), removing the similarity computation of FET$_t$ mechanism ($-$Similarity), using thresholded text mask instead of text probability (+output mask), and full model. All experiments are trained exactly the same. The qualitative results are visualized in Fig.\ref{fig3}. Since the SCUT-Syn dataset has no accurate text localization GT, we only provide the quantitative results of the six image-level evaluation metrics as in Tab.\ref{tab1}.

We observe that the performance decreases the most if removing all the FEM parts in FETNet, which demonstrates the importance of the FEM module in FETNet. It assists in erasing text features and avoids inducing them again during background reconstruction. Without FEM, the removal results in Fig.\ref{fig3}(b) contains some original text color in the erased text regions. The same problem exists for removing the FTM module in FETNet. When we remove the similarity module, there are a lot of small bits of text left over in the output. The text regions on complex background are not fixed well, as displayed on the dark green circle region in Fig.\ref{fig3}(d).

Using the 0/1 mask instead of the text confidence map, the model becomes a two-stage network, which first generates the mask and reconstructs the background through image inpainting. While using a text confidence map in FETNet makes the tasks nonindependent. The text removal-related losses can also be backpropagated to update the parameters in the text segmentation module. That results in more accurate text segmentation and better text removal results, as reflected in Fig.\ref{fig3}(f). The above results illustrate the effectiveness of each module in FETNet.

\subsubsection{The effect of various FET structures}
For evaluating the influence of the FET$_t$ and FET$_s$, we build five models with different FET structures as follows: removing all FET$_t$ ($-$FET$_t$), removing all FET$_s$($-$FET$_s$), only using FET$_t$ in all the layers (all FET$_t$), only using FET$_s$ in all the layers (all FET$_s$), and our proposed model (Full model). The quantitative results are shown in Tab.\ref{tab_structures}.

We can observe that the performance decreases more when we remove all the structure FET or the texture FET in the models compared with the performance of models using all FET$_t$ or FET$_s$ structures. It demonstrates the importance of both structure and texture FET in our method. On the other side, the model without FET$_t$ gets the worst result, but the model with all FET$_t$ gets the result only second to that of the full model. It illustrates that texture FET is more important than structural FET. Since there are convolution parameters in FET$_t$, the model size with all FET$_t$ is bigger than others. However, the FET$_s$ module has no parameters, so the -FET$_t$ model has nearly the same size as all FET$_s$ model. About the inferring time, FET$_t$ module costs more time than FET$_s$ as the similarity is computed spatially in FET$_t$. Considering the performance, model size, and speed comprehensively, our proposed full model has a good tradeoff among all these factors.





\begin{table}
  \caption{Performance of different structured models on SCUT-EnsText and SCUT-Syn.}
  \center
  \resizebox{\columnwidth}{!}{

\begin{tabular}{|cc|c|c|c|c|c|c|c|c|c|c|c|c|c|c|}
    \hline
    \multicolumn{2}{|c|}{dataset} & \multicolumn{6}{|c|}{SCUT-EnsText}& \multicolumn{6}{|c|}{SCUT-Syn} & &  \\
    \hline
    \multicolumn{2}{|c|}{Method} & PSNR $\uparrow$ & MSSIM $\uparrow$ & MSE $\downarrow$ & AGE $\downarrow$ & pEPs $\downarrow$ & pCEPs $\downarrow$  & PSNR $\uparrow$ & MSSIM $\uparrow$ & MSE $\downarrow$ & AGE $\downarrow$ & pEPs $\downarrow$ & pCEPs $\downarrow$ &Parameters $\downarrow$ &Infer. Time $\downarrow$ \\
    \hline 
     \multicolumn{2}{|c|}{- FET$_t$}  &32.47 & 95.39 & 0.0017 & 2.0661 & 0.0187 & 0.0095 &37.10 & 96.57 & 0.0003 & 1.4309 & 0.0083 & 0.0016 & \textbf{7.05M} &  \textbf{2ms}  \\
    \hline
    \multicolumn{2}{|c|}{- FET$_s$} &32.53 & 95.49 & 0.0017 & 2.0569 & 0.0185 & 0.0095 &37.15 & 96.57 & 0.0003 & 1.4157 & 0.0086 & 0.0016 &8.49M & 4.26ms \\
    \hline
     \multicolumn{2}{|c|}{all FET$_s$} &33.72 & 96.13 & 0.0016 & 1.9136 & 0.0173 & 0.0080 &37.23 & 96.65 & 0.0003 & 1.4138 & 0.0080 & 0.0015 & 7.06M & 3.39ms  \\
    \hline 
     \multicolumn{2}{|c|}{all FET$_t$}  &33.81 & 96.21 & 0.0016 & 1.8949 & 0.0171 & 0.0077  &37.52 & 96.73 & 0.0003 & 1.3931 & 0.0077 & 0.0014 & 8.72M & 4.73ms  \\
    \hline
     \multicolumn{2}{|c|}{Full model} & \textbf{34.53} & \textbf{97.01} & \textbf{0.0013} & \textbf{1.7539} & \textbf{0.0137} & \textbf{0.0080} & \textbf{39.14} & \textbf{97.97} & \textbf{0.0002} & \textbf{1.2608} & \textbf{0.0046} & \textbf{0.0008} & 8.53M & 4.62ms  \\

    \hline
\end{tabular}}
\label{tab_structures}
\end{table}

\subsection{Comparison with state-of-the-art approaches}

We compare our FETNet with the baseline fully convolutional network (FCN)  \cite{long2015fully}, three SOTA one-stage methods: Pix2pix \cite{2017Image}, STE \cite{2017Scene}, and EnsNet \cite{2018EnsNet}, and the two SOTA two-stage approach EraseNet \cite{2020EraseNet} and MTRNet++ \cite{tursun2020mtrnet++} on SCUT-EnsText, SCUT-Syn dataset, and Flickr-ST dataset. The FCN-based method is similar to STE but with the input of full image instead of image patches. It is compared as a baseline model. We retrain all the models with the original setting as officially reported but set the input image to 256$\times$256. Especially, Precision (P), Recall (R), and F-score (F) are measured on rescaled 512$\times$512 images by text detector CRAFT \cite{2019Character}, since this text detector is very sensitive to small text and cannot detect text in very small-sized images.
\begin{table}[t]
\center
  \caption{Comparison with SOTA methods and proposed method on SCUT-EnsText and Flickr-ST. R: Recall; P: Precision; F: F-score.}
    \resizebox{\linewidth}{!}{
  \begin{tabular}{|c|c|c|c|c|c|c|c|c|c|c|}
    \hline
    \multirow{2}{*}{Dateset}&\multirow{2}{*}{Method}&\multicolumn{6}{c|}{Image-level Evaluation}&\multicolumn{3}{c|}{Detection-Eval(\%)}\\\cline{3-11}
    && PSNR $\uparrow$ & MSSIM $\uparrow$ & MSE $\downarrow$ & AGE $\downarrow$ & pEPs $\downarrow$ & pCEPs $\downarrow$  &
    P $\downarrow$ & R $\downarrow$ & F $\downarrow$ \\
    \hline
    \multirow{8}{*}{SCUT-EnsText}&Original Images & -  & - & - & - & - & - & 79.8 & 69.7 & 74.4 \\ \cline{2-11}
    &FCN(baseline) & 18.16 & 47.24 & 0.0211 & 20.9734 & 0.3115 & 0.1338& 1.33& 31.37 & 2.55 \\\cline{2-11}
    &Pix2pix\cite{2017Image}  & 26.75 & 88.93 & 0.0033 & 5.842 & 0.048 & 0.0172 & 71.3 & 36.5 & 48.3 \\\cline{2-11}
    &STE\cite{2017Scene}  & 20.60 & 84.11 & 0.0233 & 14.4795 & 0.1304 & 0.0868 & 52.3 & 14.1 & 22.2 \\\cline{2-11}
    &EnsNet\cite{2018EnsNet} & 29.54 & 92.74 & 0.0024 & 4.1600 & 0.2121 & 0.0544 & 68.7 & 32.8 & 44.4 \\\cline{2-11}
    &MTRNet++\cite{tursun2020mtrnet++}  & 27.34 & 95.20 & 0.0033 & 8.9480 & 0.0953 & 0.0619 & 56.1 & \textbf{3.6} & \textbf{6.8}\\\cline{2-11}
    &EraseNet(official report)\cite{2020EraseNet}  & 32.30 & 95.42 & 0.0015 & 3.0174 &0.0160 & 0.0090 & 53.2 & 4.6 & 8.5\\\cline{2-11}
    &EraseNet(retrained)  & 31.72 & 91.72 & 0.0015 & 2.5476 & 0.0724 & 0.0508 & 65.6 & 12.9 & 21.5 \\\cline{2-11}
    &FETNet (Ours) & \textbf{34.53} & \textbf{97.01} & \textbf{0.0013} & \textbf{1.7539} & \textbf{0.0137} & \textbf{0.0080}& \textbf{51.3} & 5.8 & 10.5\\
    \hline
    \multirow{7}{*}{Flickr-ST}&Original Images & -  & - & - & - & - & - & 84.4 & 73.8 & 78.8  \\ \cline{2-11}
    &FCN(baseline)\cite{long2015fully} & 17.22 & 43.22 & 0.0237 & 22.8852 & 0.3399 & 0.1430& 2.18 &38.37 & 4.12 \\\cline{2-11}
    &Pix2pix\cite{2017Image} & 24.95 & 86.81 & 0.0059 & 7.0942 & 0.0630 & 0.0316& \textbf{45.7} & 12.8 & 20.0 \\\cline{2-11}
    &STE\cite{2017Scene} & 25.28 & 88.57 & 0.0055 & 7.6071 & 0.0569 & 0.0277& 54.0 & 7.9 & 13.8  \\\cline{2-11}
    &MTRNet++\cite{tursun2020mtrnet++}  & 26.19 & 94.37 & 0.0042 & 10.2778 & 0.1199 & 0.0777 & 71.2 & 3.3 & 6.4\\\cline{2-11}
    &EnsNet\cite{2018EnsNet} & 29.14 & 92.21 & 0.0034 & 4.1160 & 0.0311 & 0.0158 & 70.9 & 17.2 & 27.7\\\cline{2-11}
    &EraseNet\cite{2020EraseNet}  & 27.87 & 89.70 & 0.0035 & 5.4443 & 0.0311 & 0.0140& 62.3 & 5.7 & 10.4 \\\cline{2-11}
    &FETNet (Ours) & \textbf{33.61} & \textbf{95.99} & \textbf{0.0022} & \textbf{1.9542} & \textbf{0.0190} & \textbf{0.0108}& 59.0 & \textbf{2.6} & \textbf{5.0}\\
    \hline
\end{tabular}
\label{tab_real}}
\end{table}

\begin{figure}
  \includegraphics[width=\linewidth]{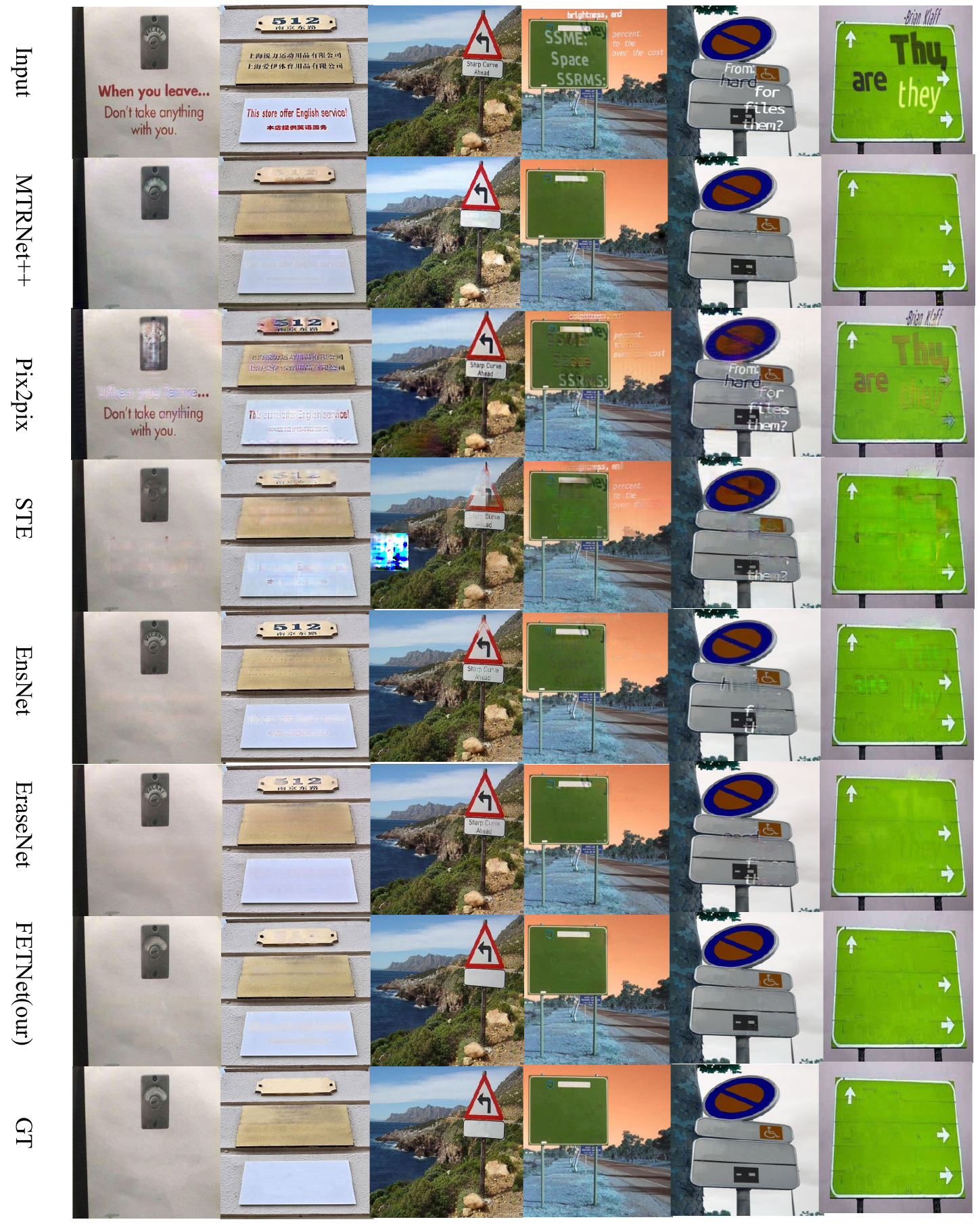}
  \caption{Comparison results with other SOTA methods on the SCUT-EnsText and SCUT-Syn datasets.}
  \label{sota_scut_real_syn}
\end{figure}

\begin{figure}
  \includegraphics[width=0.95\linewidth]{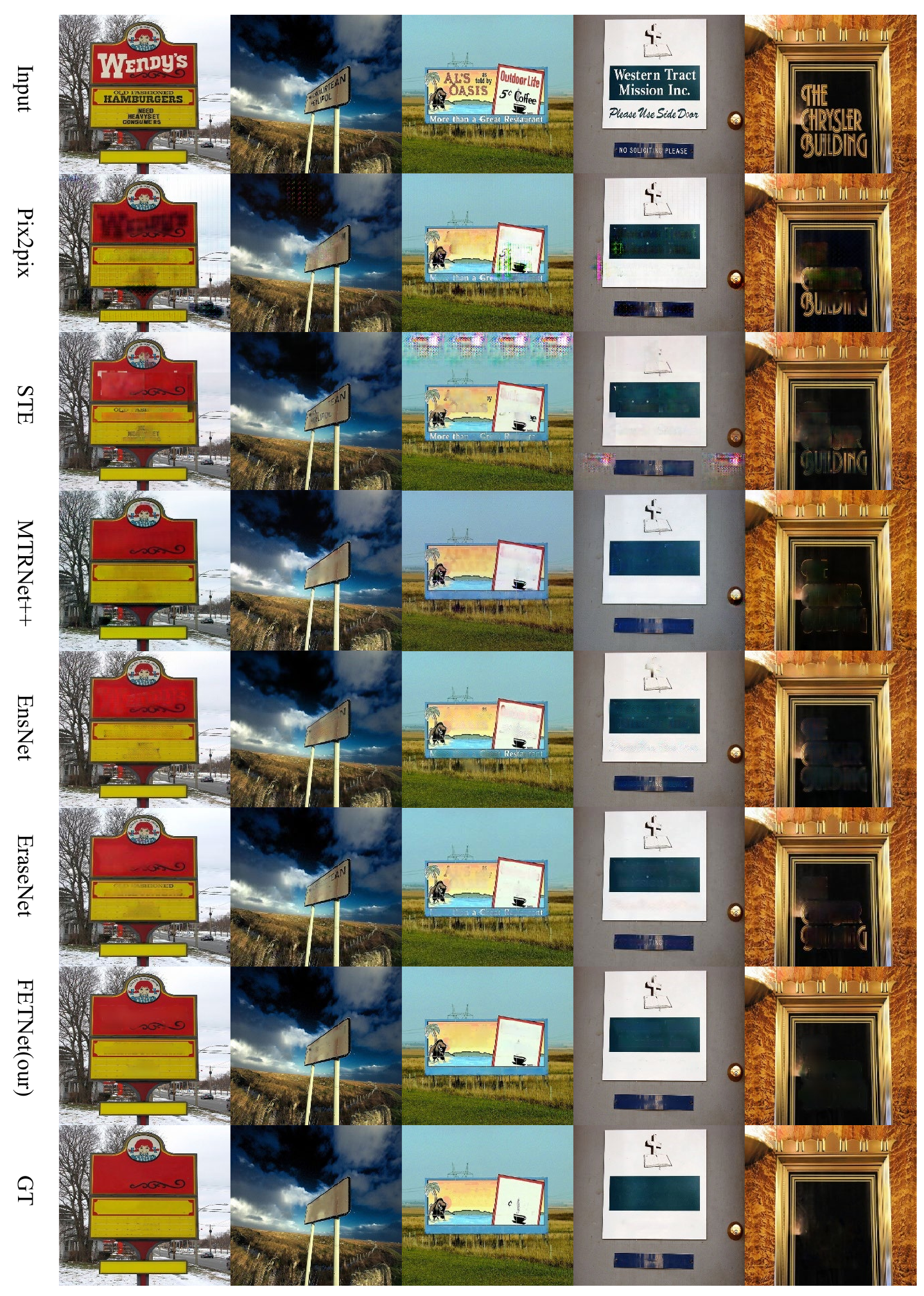}
  \caption{Comparison results with other SOTA methods on the Flickr-ST dataset.}
  \label{sota_flickr}
\end{figure}

\textbf{Qualitative Comparison}. Qualitative comparison results are displayed in Fig.\ref{sota_scut_real_syn} and \ref{sota_flickr}. Compared with one-stage STR methods, we can see that STE causes a lot of visual artifacts such as color discrepancy, blurriness, etc. Some non-text regions are also erased, reflecting the uncontrollability of its text removal process. On the contrary, our proposed method adopts an attention mechanism to reconfigure text features and utilizes text segmentation as guidance. Therefore, it can overcome these drawbacks.

Compared with other SOTA approaches on SCUT-EnsText, our method is able to remove most scene text and recover the background well. Particularly, it performs better for removing small numbers and cursive text. We can see that other methods fail to remove small text in the first and second columns of Fig.\ref{sota_scut_real_syn}, but we can correctly segment those text and remove them. The STR results of our model have significantly fewer noticeable inconsistencies and artifacts, such as color discrepancy and blurriness, as shown in the 5$_{th}$ and 6$_{th}$ columns in Fig.\ref{sota_scut_real_syn}. This indicates our model can generate more semantically plausible and elegant results.

The same results can be observed on the Flickr-ST dataset in Fig.\ref{sota_flickr}. The scene text in the Flickr-ST dataset appears to be more complicated. For instance, illumination reflection in the image of the first column, text-like background objects in the second column, and more complex background textures in the third and fourth columns. Compared to other SOTA approaches, our proposed FETNet can erase those text clearer and also recover the complex background information better.

\textbf{Quantitative Comparison}. Quantitative comparison results of real scene data and synthetic data are given in Tab.\ref{tab_real} and Tab.\ref{tab_syn}, respectively. We can see that the proposed FETNet surpasses the existing SOTA methods on most evaluation metrics on the three datasets. Additionally, FETNet is a one-stage, lightweight, and end-to-end network, which only has 8.53M trainable parameters and an inference speed of 4.62 ms per image, as indicated in Tab.\ref{tab_syn}. We can see that the one-stage STR methods run faster than the two-stage ones. They also have much smaller model sizes. The one-stage model, STE \cite{2017Scene}, generates many 1024 channel feature maps and inputs small patches of images iteratively. So, its size is large and has a high computational cost. With the guidance of the text region mask and the coarse STR results, the two-stage models are able to achieve better STR performance both on inpainting and text detection metrics, but their inferring time is slower, i.e. MTRNet++. Our model absorbs the advantages of one-stage and two-stage models. The overall comparison results demonstrate the superiority of FETNet for the STR task.

Overall, the qualitative and quantitative results indicate that our method can produce excellent text removal and background restoration results on both real scenes and synthetic text images.

\begin{table}
\center
  \caption{Comparison with SOTA methods and proposed method on SCUT-Syn.}
  \resizebox{\linewidth}{!}{
  \begin{tabular}{|c|c|c|c|c|c|c|c|c|c|}
    \hline
    Method &Category& PSNR $\uparrow$ & MSSIM $\uparrow$ & MSE $\downarrow$ & AGE $\downarrow$ & pEPs $\downarrow$ & pCEPs $\downarrow$&Parameters$\downarrow$&Infer. Time$\downarrow$\\
    \hline
    FCN(Baseline)\cite{long2015fully} &one-stage & 18.98 & 53.27 & 0.0196 & 19.3350 & 0.2777 & 0.1225 & 18.13M  &\textbf{2.23ms}\\
    \hline
    Pix2pix &one-stage & 25.16 & 87.63 & 0.0038 & 6.8725 & 0.0664 & 0.0300 & 54.4M &2.96ms \\
    \hline
    STE &one-stage & 24.02 & 89.49 & 0.0123 & 10.0018 & 0.0728 & 0.0464 & 89.16M&18.45ms \\
    \hline
    EnsNet &one-stage & 37.36 & 96.44 & 0.0021 & 1.73 & 0.0069 & 0.0020 & 12.4M &5.10ms \\
    \hline
    MTRNet++ (official  report) &two-stage & 34.55 & 98.45 & 0.0004 & - & - & - & 18.7M & - \\
    \hline
    MTRNet++ (retrained) &two-stage & 34.28 & 96.78 & 0.0004 & 2.6337 & 0.0059 & 0.0008 & 18.7M & 30.2ms \\
    \hline
    EraseNet(official  report) &two-stage & 38.32 & 97.67 & \textbf{0.0002} & 1.5982 & 0.0048 & \textbf{0.0004} & 19.74M &8.67ms\\
    \hline
    EraseNet(retrained) &two-stage & 37.43 & 97.57 & 0.0003 & 1.5615 & 0.0264 & 0.0140 & 19.74M &8.67ms\\
    \hline
    FETNet(Ours) &one-stage & \textbf{39.14} & \textbf{97.97} & \textbf{0.0002} & \textbf{1.2608} & \textbf{0.0046} & 0.0008 & \textbf{8.53M} &4.62ms\\
    \hline
\end{tabular}
\label{tab_syn}}
\end{table}

\subsection{Inserting FET for two-stage network}
The FET mechanism can be used as a drop-in module inserted in the skip-connection structure for other two-stage STR networks, which utilize text masks and are built on encoder-decoder architecture. We next investigate the effect of inserting FET in the coarse detection stage of EraseNet. The FET$_t$ blocks are inserted in the first three CNN layers and FET$_t$ in the last two layers as our FETNet design. Both the original EraseNet and FET-inserted one are trained in the same condition on the SCUT-Syn dataset. 

The test results are given in Tab.\ref{tab4}. By inserting our FET blocks, the EraseNet can get higher PSNR, MSSIM, and lower MSE, AGE, pEPs, pCEPs. The better performance demonstrates the effectiveness of our proposed FET mechanism.

\begin{table}
\center
  \caption{The performance of inserting FET to EraseNet on SCUT-Syn.}
  \resizebox{0.72\linewidth}{!}{
  \begin{tabular}{|c|c|c|c|c|c|c|}
    \hline
    Method y& PSNR $\uparrow$ & MSSIM $\uparrow$ & MSE $\downarrow$ & AGE $\downarrow$ & pEPs $\downarrow$ & pCEPs $\downarrow$\\
    \hline
    EraseNet  & 37.43 & 97.57 & 0.0003 & 1.5615 & 0.0264 & 0.0140   \\
    \hline
    FET+EraseNet  & 38.15 & 98.01 & 0.0002 & 1.0865 & 0.0124 & 0.0060  \\
    \hline
\end{tabular}}
\label{tab4}
\end{table}

\subsection{Interpretability}

\begin{figure}
\center
  \includegraphics[height=0.4\textwidth]{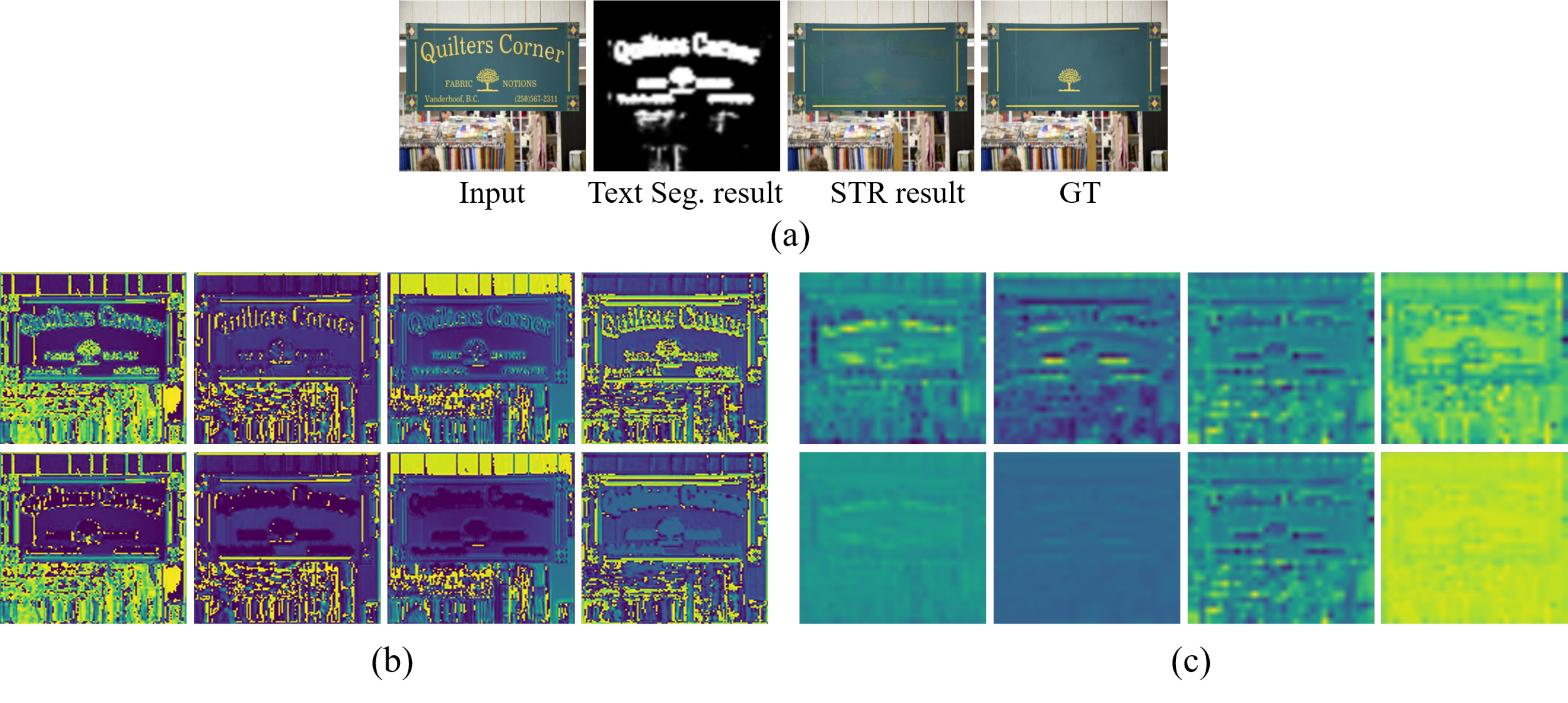}
  \caption{Visualizations of features revised by the FET mechanism. (a) lists the input image, generated mask, STR results, and GT. (b) Features input(1st row) and output(2nd row) of FET$_t$. (c) Features input(1st row) and output(2nd row) of FET$_s$.}
  \label{visual}
\end{figure}

To further understand the working mechanism of FET, we observe the original CNN features in the second layer and fourth layer of the encoder and their corresponding output features after using FET$_t$ and FET$_s$ respectively. The visualization results of example texture and structure features are displayed in Fig.\ref{visual}(b) and (c) by using the input image in (a). We can see that highlighted features in text regions are removed and can be refilled in by the surrounding background features, as displayed in Fig.\ref{visual}(b), which proves the effectiveness of our proposed FET$_t$.

FET$_s$ revises features based on channel attention. If the original channel feature contains text structure, it is decreased after using FET$_s$ as shown in the first two columns of Fig.\ref{visual}(c). While if the original channel contains important background structures, the corresponding channel features output from FET$_s$ are maintained or enhanced, as shown in the last two columns of Fig.\ref{visual}(c). These observations illustrate that FET$_s$ is able to select background structure but suppress text structure features channel-wisely.






\section{Limitation}

\begin{figure}
  \centering
  \includegraphics[height=0.5\textwidth]{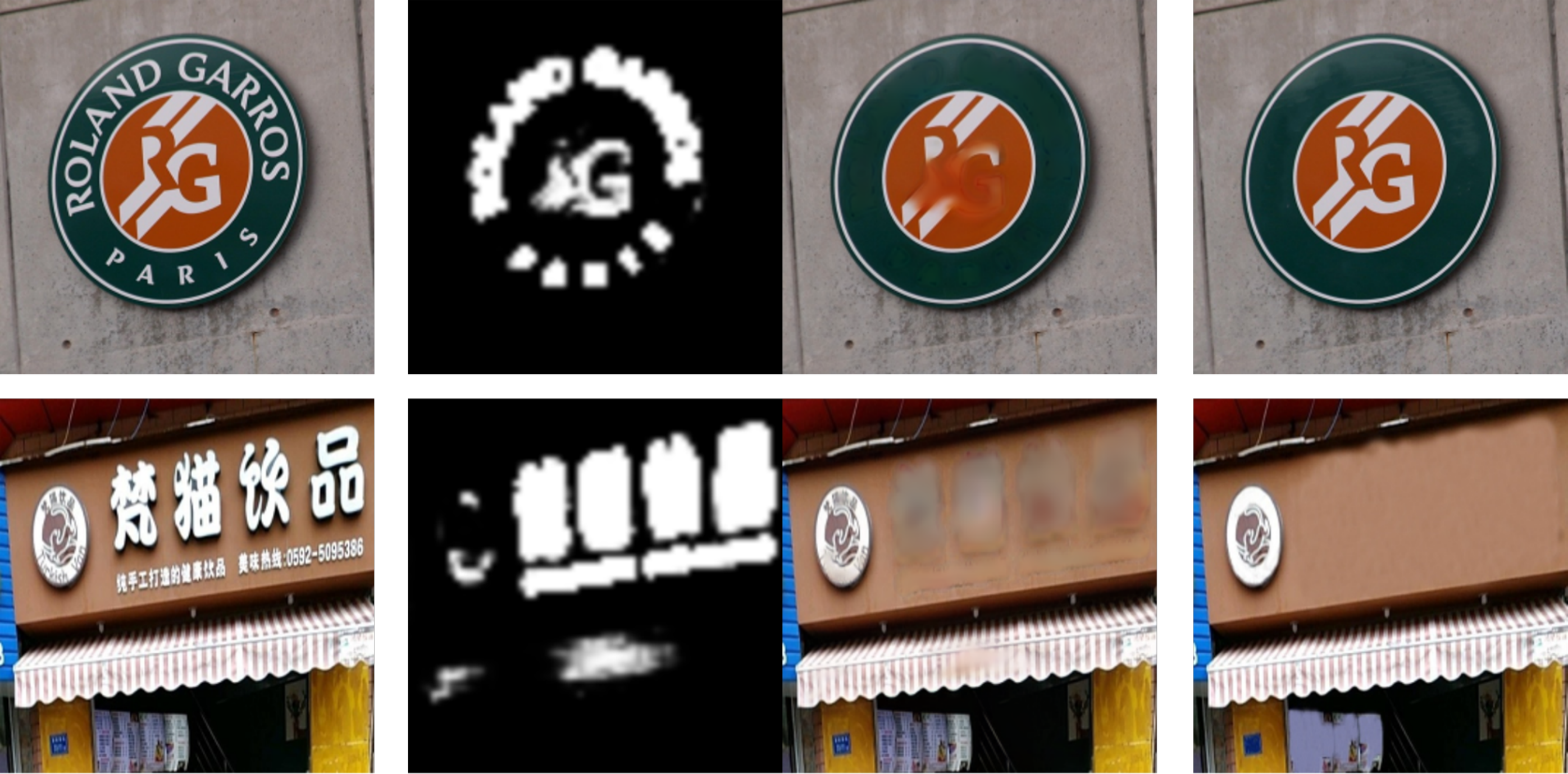}
   \caption{Some failure cases of our method for STR.}
   \label{lm}
\end{figure}

Even though our method achieves great performance on scene text removal tasks, it still has some limitations. If the text blends in the background, as shown in the top row of Fig.\ref{lm}, where the central characters ``R'' and ``G'' are formed by the background textures, it will be difficult to distinguish them between text and non-text. The incorrect segmented text map results in limited text removal performance. When the text has complex text effects, e.g., shadows, the removal results are also unsatisfactory. For failure example in the bottom of Fig.\ref{lm}, the large-size text is stereoscopic and has shadows, making the incorrect segmentation of text. The components of segmented text are very big. However, filling in the large holes in the image inpainting task is very challenging~\cite{2020Recurrent}. Therefore, our model still has room for improvement in more accurate artistic text extraction and considering using image inpainting strategies for filling in large holes.

\section{Conclusion}

This paper presents a lightweight one-stage network, called FETNet, for scene text removal in images. It is based on an encoder-decoder structure with inserted FET blocks. The FET block contains three modules: FEM, SAM/CAM, and FTM, for text feature erasing, background feature exploring, and transferring correct background features to text regions, respectively, with the text probability guidance generated from the text segmentation branch. Therefore, the decoder can generate better text removal results using these reconfigured features. The experiments show that our FETNet significantly outperforms the existing state-of-the-art STR methods. Furthermore, we show that the FET mechanism can be inserted into other two-stage methods for performance enhancement. Additionally, we modified the Flickr-ST dataset, which can benefit multiple STR-related researches, e.g., Scene text detection, segmentation, recognition, and removal. In the future, we will fuse the knowledge of natural language processing into our work to better remove scene text with private information in the images.

\section{Acknowledgement}

This work is supported by the Open Project Program of the National Laboratory of Pattern Recognition (NLPR) (No.202200049).





\bibliographystyle{elsarticle-num}
\bibliography{egbib}




\end{document}